\documentclass[runningheads]{llncs}
\usepackage{color,xcolor}
\usepackage{latexsym}
\usepackage{multirow}
\usepackage{subfigure}
\usepackage{url}
\usepackage{graphicx}
\usepackage{array}
\usepackage{enumitem}
\usepackage{ulem}
\usepackage{booktabs}
\usepackage{amssymb}
\usepackage{marvosym}
\usepackage{ifsym}
\usepackage{mathrsfs}
\usepackage{amsmath}
\newcommand{\tabincell}[2]{\begin{tabular}{@{}#1@{}}#2\end{tabular}} 

\begin{document}
\title{Incorporating Commonsense Knowledge into Story Ending Generation via Heterogeneous Graph Networks}
%
\titlerunning{Commonsense Knowledge Story Ending Generation}
%
\author{
Jiaan Wang\inst{1}\thanks{indicates equal contribution.} \and
Beiqi Zou\inst{3\star} \and
Zhixu Li\inst{2} \and
Jianfeng Qu\inst{1(\textrm{\Letter})} \and
\\
Pengpeng Zhao\inst{1} \and
An Liu\inst{1} \and
Lei Zhao\inst{1}
}
\authorrunning{J. Wang et al.}
\institute{School of Computer Science and Technology, Soochow University, Suzhou, China \and
Shanghai Key Laboratory of Data Science, School of Computer Science, Fudan University, Shanghai, China \and
Department of Computer Science, Princeton University, USA \\
\email{jawang1@stu.suda.edu.cn, bzou@cs.princeton.edu}
\email{zhixuli@fudan.edu.cn, \{jfqu,ppzhao,anliu,zhaol\}@suda.edu.cn}
}
\maketitle              
\begin{abstract}
Story ending generation is an interesting and challenging task, which aims to generate a coherent and reasonable ending given a story context.
The key challenges of the task lie in how to comprehend the story context sufficiently and handle the implicit knowledge behind story clues effectively, which are still under-explored by previous work.
In this paper, we propose a Story Heterogeneous Graph Network (SHGN) to explicitly model both the information of story context at different granularity levels and the multi-grained interactive relations among them.
In detail, we consider commonsense knowledge, words and sentences as three types of nodes. To aggregate non-local information, a global node is also introduced.
Given this heterogeneous graph network, the node representations are updated through graph propagation, which adequately utilizes commonsense knowledge to facilitate story comprehension.
Moreover, we design two auxiliary tasks to implicitly capture the sentiment trend and key events lie in the context. The auxiliary tasks are jointly optimized with the primary story ending generation task in a multi-task learning strategy.
Extensive experiments on the ROCStories Corpus show that the developed model achieves new state-of-the-art performances. Human study further demonstrates that our model generates more reasonable story endings.


\end{abstract}

\section{Introduction}

\begin{figure}[t]
\centerline{\includegraphics[width=0.80\textwidth]{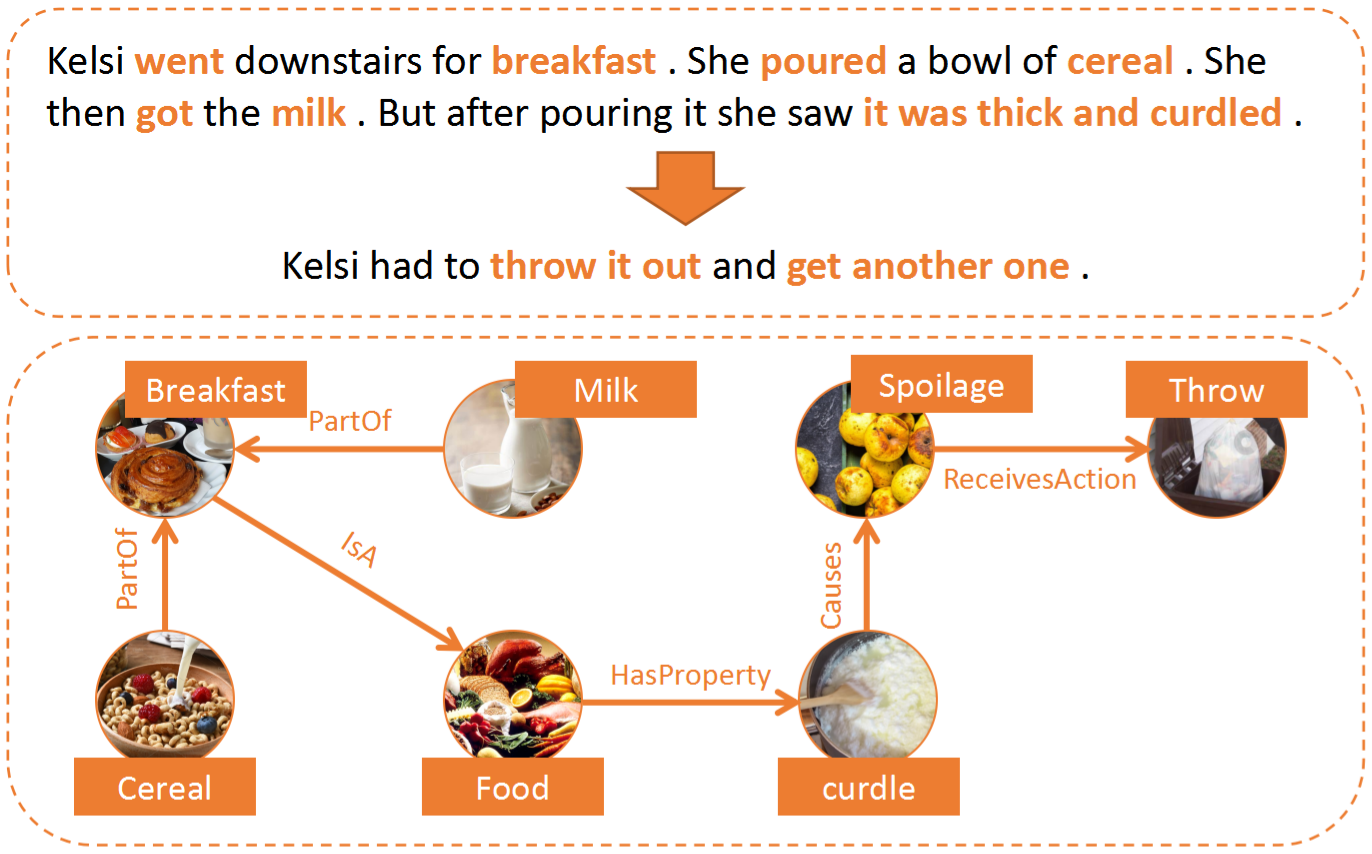}}
\caption{ The top graph shows an example of story ending generation. Orange words are entities and events. The bottom graph indicates the implicit knowledge behind the story.}
\label{fig:example}
\end{figure}

Story ending generation (SEG) is a natural language generation task, which aims at concluding a story ending given a context~\cite{Zhao2018FromPT}.
Generally, a story context contains a series of entities and events (known as story clues), each of which could have strong logical relationships with others, which leads to rich interactive relations across the whole context.
For humans, one may utilize his/her own commonsense knowledge to capture the story clues and conceive story endings.
As shown in Figure~\ref{fig:example}, the story clue of example story is:  \textit{went}\_\textit{breakfast} $\Rightarrow$ \textit{poured}\_\textit{cereal} $\Rightarrow$ \textit{got}\_\textit{milk} $\Rightarrow$ \textit{it}\_\textit{was}\_\textit{thick}\_\textit{and}\_\textit{curdled}, which indicates the food may have gone bad. It is natural for humans to \textit{throw the bad food and get another one} since we all know \textit{spoiled food is harmful to health}.
Therefore, in order to generate a coherent and reasonable ending, generative models should not only sufficiently comprehend the story context and further capture the story clues but also effectively handle the implicit knowledge behind them.

Most previous work views the story context as a linear sequence of words and ignores the rich relations among them. For example, Zhao et al.~\cite{Zhao2018FromPT} and Li et al.~\cite{li-etal-2018-generating} explore variant sequence-to-sequence (Seq2Seq) models encoding the story context in a left-to-right manner and decoding endings. They further utilize reinforcement learning to improve the rationality and/or diversity of generated endings.
Gupta et al.~\cite{gupta-etal-2019-writerforcing} resort to an extra keywords extraction algorithm and model the keywords information in the Seq2Seq models. Luo et al.~\cite{luo-etal-2019-learning} make use of existing sentiment analyzers to consider the fine-grained sentiment in their Seq2Seq method.
Guan et al.~\cite{Guan2019StoryEG} design incremental encoding and multi-source attention mechanism to model the relation of adjacent sentences and incorporate commonsense knowledge into contextual representation.
These methods all adopt sequential modeling strategies for encoding the story context, hindering the exploration of inherently rich interactive relations across the whole context, which makes the story context and commonsense knowledge modeling inadequate.

Recently, Huang et al.~\cite{Huang2021StoryEG} suggest that the great importance of story clues hidden in the context and further propose multi-level graph convolutional networks over dependency parse (MGCN-DP) that models SEG task in a graph-to-text manner.
The graph architecture better models the interactive relations across story context and result in more coherent endings compared to sequential methods.
However, the relations of words from different context sentences cannot be explicitly captured by MGCN-DP which only contains word nodes from the same sentence in a graph. Besides, the MGCN-DP model does not consider commonsense knowledge behind the story, thus the generated endings could be suboptimal.

To remedy above issues, in this paper, we propose a \textbf{S}tory \textbf{H}eterogeneous \textbf{G}raph \textbf{N}etwork (SHGN) for SEG. The heterogeneous graph network shows its superiority in many tasks, such as recommender systems and summarization~\cite{Feng2021DialogueDG,Feng2021IncorporatingCK}.
Specifically, three types of graph nodes are considered in our SHGN: \textit{commonsense knowledge}, \textit{words} and \textit{sentences}.
Besides, a \textit{global} node is also introduced to the graph to aggregate the non-local information (see Figure~\ref{fig:graph_construct}).
To obtain contextualized representations for these nodes, large-scale pre-trained embedding models such as Sentence-BERT~\cite{reimers-gurevych-2019-sentence} and SimCSE~\cite{Gao2021SimCSESC} are used for contextual encoding.
Then, a graph neural network is used to propagate message and update representations of nodes in the heterogeneous graph.
The final node representations are passed through transformer decoders to generate story endings.
In addition, to sufficiently comprehend the story context, we design two sub-tasks with special consideration for story ending generation:
(1) the \textit{sentiment prediction of story endings sub-task} uses representations of sentence nodes to predict the sentiment of corresponding story ending, which is constructed to push the model to capture the fine-grained sentiment trend;
(2) the \textit{clue words prediction sub-task} utilizes representations of word nodes to predict whether each word belongs to story clues, which is expected to force the model to identify the key events in the context.
Such two sub-tasks can be coupled with the primary story ending generation task via multi-task learning strategy, resulting in our final model SHGN.

We conduct various experiments on the widely used ROCStories Corpus~\cite{Mostafazadeh2016ACA}. Experimental results show that our approach achieves state-of-the-art performances on SEG task. Human study indicates that our SHGN generates more coherent and reasonable story endings as compared to previous strong baselines.

\setcounter{footnote}{0}
Our main contributions in this paper are summarized as follows:  
\begin{itemize}
\item We propose a Story Heterogeneous Graph Network for SEG, which explicitly models the information of story context at different granularity levels and the multi-grained interactive relations among them\footnote{We release our code and generated results at \url{https://github.com/krystalan/AwesomeSEG}.}. 
\item We also design two auxiliary tasks (i.e., sentiment prediction of story endings and clue words prediction) to facilitate story comprehension. To the best of our knowledge, we are the first to apply multi-task learning strategy on SEG.
\item Extensive experiments on widely used ROCStories Corpus show that our model achieves new state-of-the-art performances. Human study and case study further prove that our model could generate more coherent and reasonable story endings.
\end{itemize}

\section{Related Work}
\label{sec:related_work}

\noindent \textbf{Story Generation.} Story Generation (SG), also known as storytelling, aims at generating a logical self-consistent story plot. Early SG work~\cite{Gervs2005StoryPG,Riedl2010NarrativePB} mainly uses case-based or planning-based methods.
Recently, researchers focus on generating stories with storyline or intermediate representations, such as skeletons~\cite{xu-etal-2018-skeleton}, events~\cite{Martin2018EventRF}, titles~\cite{Li2019LearningTW} and verbs~\cite{Tambwekar2019ControllableNS}. In this way, they first generate intermediate representations, then rewrite and enrich them to obtain complete stories.

\noindent \textbf{Story Ending Generation.} Story Ending Generation (SEG) is a subtask of SG, which aims to understand the context and generate a coherent and reasonable story ending~\cite{Zhao2018FromPT}. Li et al.~\cite{li-etal-2018-generating} introduce Seq2Seq model with adversarial training to improve the rationality and diversity of the generated story endings. Similarly, Zhao et al.~\cite{Zhao2018FromPT} employ Seq2Seq model based on reinforcement learning to generate more sensible endings. Gupta et al.~\cite{gupta-etal-2019-writerforcing} utilize an extra keywords extraction algorithm and model the keywords information in the proposed Seq2Seq model. Guan et al.~\cite{Guan2019StoryEG} introduce a model which uses an incremental encoding scheme and commonsense knowledge to generate reasonable endings. Further, Huang et al.~\cite{Huang2021StoryEG} propose a multi-level graph convolutional network to capture the dependency relations of input sentences.
%
Although great progress has been made, the implicit knowledge and multi-grained interactive relations behind story context are still under-explored.

\begin{figure}[t]
\centerline{\includegraphics[width=0.95\textwidth]{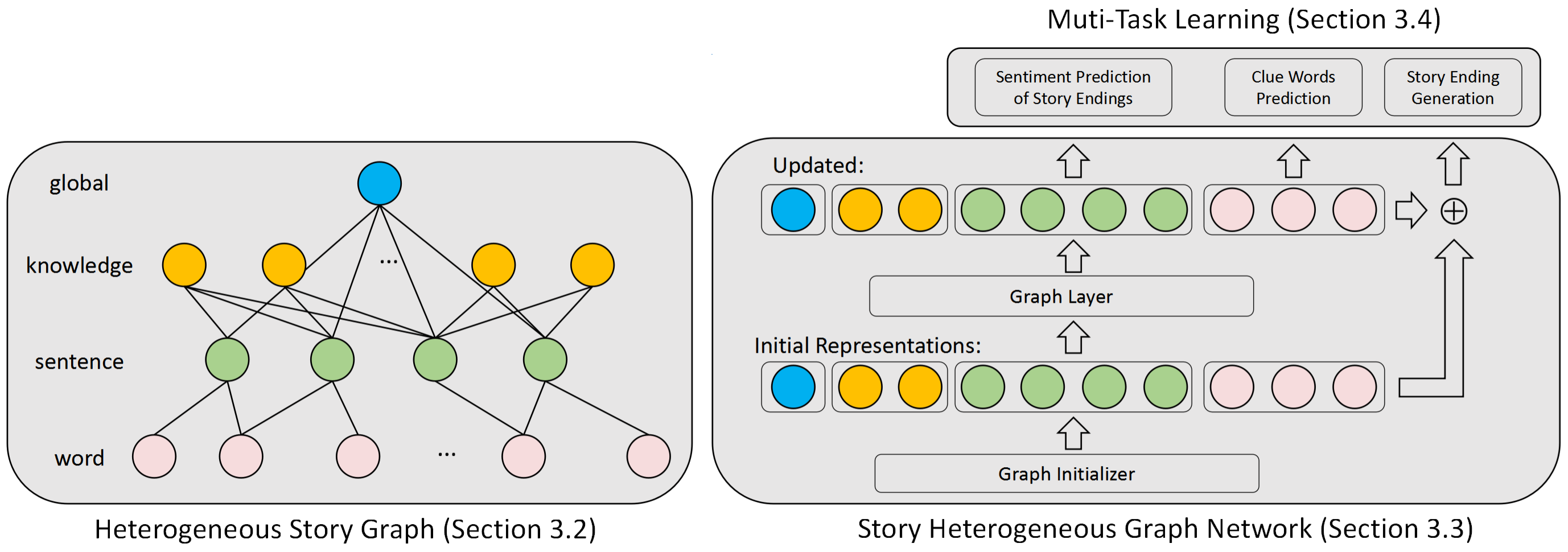}}
\caption{Overview of our proposed model. For a story context, we first construct a heterogeneous story graph (Section~\ref{subsec:graph_construct}). Then, we design our story heterogeneous graph network (SHGN) to initialize and update node representations in the graph (Section~\ref{subsec:SHGN}). Finally, auxiliary tasks are introduced to facilitate story comprehension, which are jointly optimized with the primary SEG task in the multi-task learning strategy (Section~\ref{subsec:multitask}).}
\label{fig:overview}
\end{figure} 

\section{Model}
\label{sec:model}

\subsection{Overview}
Story ending generation (SEG) task aims to generate a story ending conforming the corresponding context. Given a story context $X = \{X^{1}, X^{2}, ..., X^{\mu}\}$, where $X^{k}=x_{1}^{k}x_{2}^{k}...x_{l}^{k}\:(0 \le k \le \mu)$ represents the $k$-th sentence with $l$ words. SEG aims at generating a story ending $E=y_{1}y_{2}...y_{m}$ with $m$ words.

Figure~\ref{fig:overview} shows the overview of our proposed model.
To generate a coherent and reasonable story ending, we model the story context into a heterogeneous graph (\S~\ref{subsec:graph_construct}). Based on the graph, we propose a Story Heterogeneous Graph Network (SHGN), which contains three components: (1) \textit{graph initializer} is used to give each node an initial representation; (2) \textit{graph layer} digests the structural information and gets updated node representations; (3) \textit{transformer decoder} is used to generate the story endings according to final node representations (\S~\ref{subsec:SHGN}). Moreover, in order to sufficiently comprehend the story context, we design two auxiliary tasks, namely \textit{sentiment prediction of story endings} and \textit{clue words prediction}, which are expected to implicitly capture the sentiment trend as well as key events lie in context. These auxiliary tasks are jointly optimized with the primary story ending generation task in the multi-task learning strategy (\S~\ref{subsec:multitask}).

\begin{figure}[t]
\centerline{\includegraphics[width=0.90\textwidth]{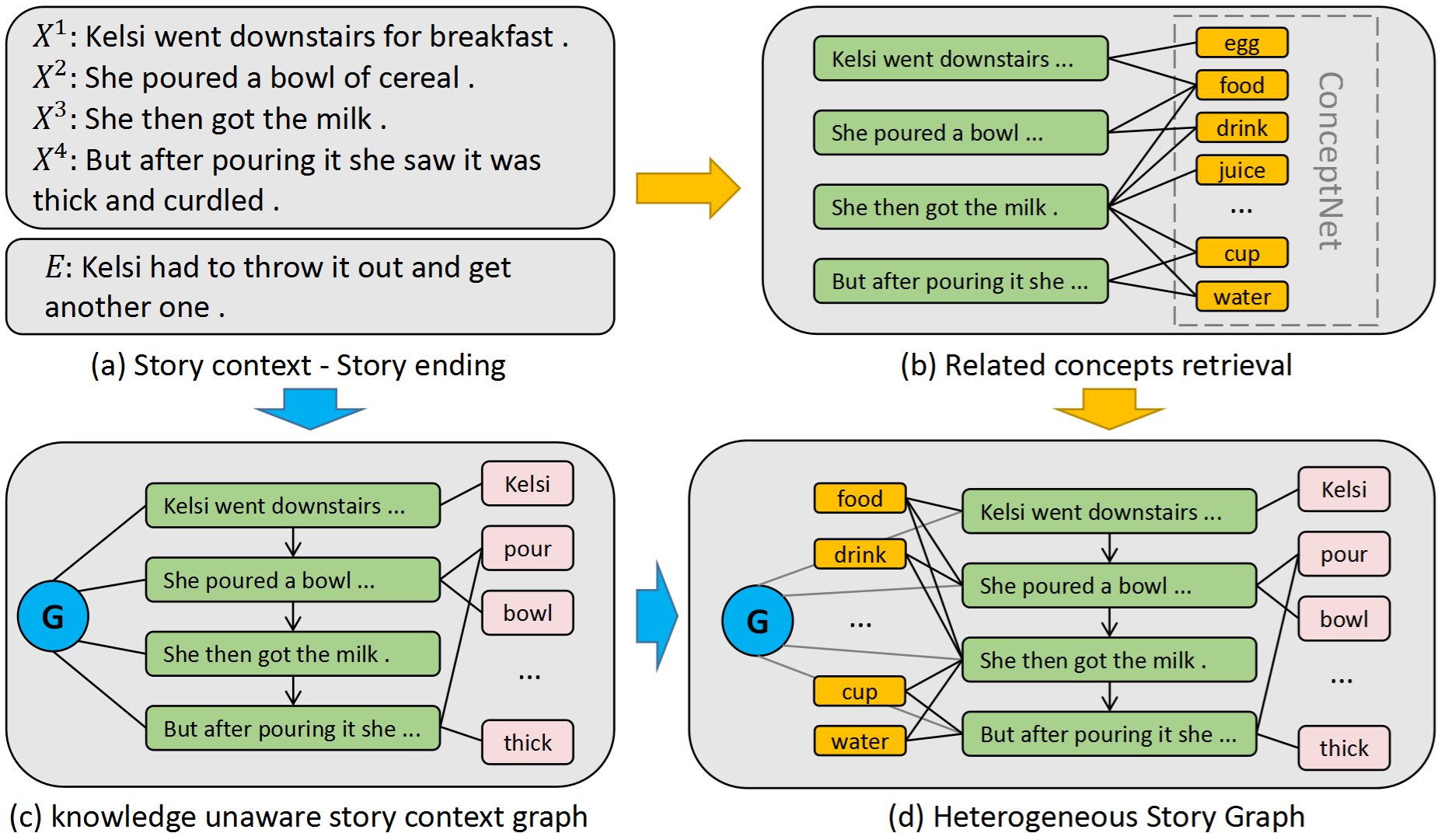}}
\caption{Illustration of heterogeneous story graph construction process. Best viewed in color. \textcolor[RGB]{0,176,240}{Blue}, \textcolor[RGB]{255,192,0}{orange}, \textcolor[RGB]{169,209,142}{green} and \textcolor{pink}{pink} colors represent global, knowledge, sentence and word nodes, respectively.}
\label{fig:graph_construct}
\end{figure} 

\subsection{Heterogeneous Story Graph Construction}
\label{subsec:graph_construct}

We define the heterogeneous story graph as a directed graph $G=(\mathcal{V}, \mathcal{E}, \mathcal{A}, \mathcal{R})$, where $v \in \mathcal{V}$ represents each node and $e \in \mathcal{E}$ denotes each edge. $\mathcal{A}$ and $\mathcal{R}$ are the sets of node types and edge types, respectively.
$\tau(v): \mathcal{V} \rightarrow \mathcal{A}$ and $\phi(e): \mathcal{E} \rightarrow \mathcal{R}$ are type mapping functions which link nodes and edges to their specific type.

Figure~\ref{fig:graph_construct} shows the construction process of heterogeneous story graph. For a given story context $X$, we utilize each word $x_{t}^{k}$ of each context sentence $X^{k}$ to retrieve related one-hop concepts from \texttt{ConceptNet} commonsense knowledge graph\footnote{We only consider nouns, verbs, adjectives, and adverbs to retrieve related concepts from \texttt{ConceptNet}.}~\cite{Speer2012RepresentingGR} (Cf. Figure~\ref{fig:graph_construct}(b)).
To model the story context, we first construct knowledge unaware story graph by viewing sentences and words as different types of nodes. As shown in Figure~\ref{fig:graph_construct}(c), each sentence node connects to both the next sentence (if has) and its word nodes (except stopwords). We also introduce a global node to aggregate non-local information. The global node connects to each sentence node using edges in both directions. Then, we combine the related concepts and knowledge unaware story graph as our heterogeneous story graph.
Specifically, we only retain concepts retrieved from more than one context sentence to control the quality of retrieved concepts.
The related concepts are regarded as knowledge nodes. If there are multiple identical knowledge nodes or word nodes, we also combine them into a single one.
Figure~\ref{fig:graph_construct}(d) shows the overview of our final heterogeneous story graph which contains four types of nodes (global, knowledge, sentence, word) and seven types of edges (global $\Rightarrow$ sentence, sentence $\Rightarrow$ global, sentence $\Rightarrow$ sentence, knowledge $\Rightarrow$ sentence, sentence $\Rightarrow$ knowledge, word $\Rightarrow$ sentence, sentence $\Rightarrow$ word) in total.

\subsection{Story Heterogeneous Graph Network}
\label{subsec:SHGN}

The Story Heterogeneous Graph Network (SHGN) is used to initialize and update each node representation in the constructed graph (\S~\ref{subsec:graph_construct}). Three components are introduced in SHGN: \textit{graph initializer}, \textit{graph layer} and \textit{transformer decoder}.

\noindent \textbf{Graph Initializer.}
The role of graph initializer is to give each node $v_{i} \in \mathcal{V}$ an initial representation $h_{v_{i}}^{0}$. For the global node, we randomly initialize its representation.
For other nodes, the initial representations should contain the information of the corresponding documents, sentences, words or concepts.
Owing to the emergence of large-scale pre-trained sentence embedding models~\cite{reimers-gurevych-2019-sentence,Gao2021SimCSESC,Huang2021WhiteningBERTAE}, which show their superiority in many sentence-level NLP tasks, we decide to utilize SimCSE~\cite{Gao2021SimCSESC} (we also try different graph initializers in the experiments, and the SimCSE performs best. More details please refer to Section~\ref{sec:ablations}) to initialize representations. It is worth noting that the SimCSE model is also used to initialize knowledge and word nodes, instead of using geometric embedding methods (e.g., TransE) or word embedding methods (e.g., GloVe). In this way, there is no need to bridge the representation gap between different node types.

\noindent \textbf{Graph Layer.} 
Given a constructed graph $\mathcal{G}$ with node representations, we use Heterogeneous Graph Transformer (HGT)~\cite{Hu2020HeterogeneousGT} as our graph layer, which models the heterogeneous graph by type-dependent parameters and can be easily applied to our heterogeneous story graph.
Specifically, HGT includes: 

(a) \textit{Heterogeneous Mutual Attention} is used to calculate the attention scores between source node $s$ and target node $t$ with the consideration of their edge $e = (s,t)$: 
\begin{equation}
Attn^{(l)}(s,e,t) = \mathop{softmax}\limits_{\forall s \in N(t)} (\alpha^{(l)}(s,e,t))
\end{equation}
\begin{equation}
\alpha^{(l)}(s,e,t) = (k_{s}^{(l)}W_{att,\phi(e)}^{(l)}q_{t}^{(l)^\top})
\end{equation}
\begin{equation}
k_{s}^{(l)} = W_{k,\tau(s)}^{(l)} (h_{s}^{(l-1)})
\end{equation}
\begin{equation}
q_{t}^{(l)} = W_{q,\tau(t)}^{(l)} (h_{t}^{(l-1)})
\end{equation}
where $N(t)$ denotes neighbors of target node $t$.  $k_{v}^{(l)}$ / $q_{v}^{(l)}$ represents the $l$-th layer key / query vector of node $v$. $h_{v}^{(l-1)}$ is the $(l$-$1)$-th layer representation of node $v$. $W_{att,\phi(e)}^{(l)}$, $W_{k,\tau(s)}^{(l)}$ and $W_{q,\tau(t)}^{(l)}$ are trainable parameters.

(b) \textit{Heterogeneous Message Passing} is utilized to pass information from source nodes to target nodes:
\begin{equation}
Message(s,e,t) = \mathop{||}\limits_{i \in [1,h]} MSGHead^{i}(s,e,t)
\end{equation}
\begin{equation}
MSGHead^{i}(s,e,t) = W_{i,\tau(s)}^{(l)} (h_{s}^{(l-1)}) W_{\phi(e)}^{MSG}
\end{equation}
where $h$ is the number of heads in HGT, $W_{i,\tau(s)}^{(l)}$ and $W_{\phi(e)}^{MSG}$ are trainable parameters.

(c) \textit{Target-Specific Aggregation} is used to aggregate information from the source nodes to the target node:
\begin{equation}
\tilde{h}_{t}^{(l)} = \mathop{\oplus}\limits_{\forall s \in N(t)} (Attn^{(l)}(s,e,t) \cdot Message(s,e,t))  
\end{equation}
\begin{equation}
h_{t}^{(l)} = W_{\tau(t)}(\sigma(\tilde{h}_{t}^{(l)})) + h_{t}^{(l-1)}
\end{equation}
where $\oplus$ and $\sigma$ represent addition operator and sigmoid function, respectively. $W_{\tau(t)}$ is trainable parameters.

After obtaining the output representation $h_{v}^{L}$ for each node, we concatenate updated node representation $h_{v}^{L}$ with corresponding initial representation $h_{v}^{0}$ and followed by a linear projection function to get the final node representation:
\begin{equation}
h_{v}^{L} = W_{final}[h_{v}^{L},h_{v}^{0}] 
\end{equation}

\noindent \textbf{Transformer Decoder.}
We utilize the vanilla Transformer decoder~\cite{Vaswani2017AttentionIA} to decode the story endings. The inputs of multi-head attention in transformer decoder are node representations $H^{L} = [h_{v_{0}}^{L};h_{v_{1}}^{L};...;h_{v_{s}}^{L}]$ ($s$ denotes the total number of nodes in the graph) and decoder input $D_{in}$. This process is denoted as:
\begin{equation}
\tilde{D}_{in} = MultiHead(D_{in},H^{L},H^{L})
\end{equation}
\begin{equation}
D_{o} = FFN(\tilde{D}_{in})
\end{equation}
where $FFN$ is two linear projections with a ReLU activation in the middle. $D_{o}$ is the middle output of transformer decoder.

To generate next word, a linear projection and softmax function are used to predict the word probabilities:
\begin{equation}
\mathcal{P}(y_{t}|y<t,X) = softmax(W_{o}D_{o}) 
\end{equation}
where $W_{o}$ is trainable parameters, $\mathcal{P}(y_{t})$ is the probability distribution over vocabulary.

Then, we calculate the negative data likelihood as loss function:
\begin{equation}
\mathcal{L}_{gen} = - \sum_{t} log \mathcal{P}(y_{t}=\tilde{y_{t}}|y<t,X)
\end{equation}

\subsection{Auxiliary Tasks}
\label{subsec:multitask}
To sufficiently comprehend the story context and generate more coherent story endings, we design two auxiliary tasks with the special consideration for SEG: \textit{sentiment prediction of story endings} and \textit{clue words prediction}.
These two auxiliary tasks are jointly optimized with the SEG task in a multi-task learning strategy.

\noindent \textbf{Sentiment Prediction of Story Endings.} Generally, the sentimental trend of the story context plays a crucial role in the SEG~\cite{Mo2021IncorporatingST}. In order to improve the sentimental consistency of the generated endings. We make use of representations of all sentence nodes to predict the sentiment of the corresponding ending:

\begin{equation}
Y_{s} = W_{s} ( \sum_{v_{s}} h_{v_{s}}^{L} )
\end{equation}
where $v_{s}$ denotes the sentence nodes and $Y_{s}$ is the sentimental probability distribution. 

We use the \texttt{VADER} toolkit~\cite{Hutto2014VADERAP} to construct sentiment labels of the gold story endings. The labels include ``positive'', ``neutral'' and ``negative''. During training, the cross entropy loss function can be defined as:
\begin{equation}
\mathcal{L}_{sen} = CE(Y_{s},\tilde{Y}_{s})
\end{equation}

\noindent \textbf{Clue Words Prediction.} The story clues that lie in context are also important for SEG. Huang et al.~\cite{Huang2021StoryEG} find that the words of top-2-degree in the dependency tree of each sentence are similar to the story clues summarized by humans. In detail, the dependency tree model the word sequences in the directed graph architecture, whose edges represent the dependency relations between words, such as causal relation, modifier relation, etc. The words of top-2-degree have the most dependency relations with others.
Thus, we utilize Biaffine~\cite{Dozat2017DeepBA} to construct the dependency tree for each sentence in the context and further regard the words of top-2-degree as clue words.
We use the representation of each word node to predict whether it is a clue word.
The binary cross entropy loss function of \textit{clue words prediction} is denoted by:
\begin{equation}
\mathcal{L}_{clu} = CE(Y_{w},\tilde{Y}_{w})
\end{equation}
\begin{equation}
Y_{w} = W_{w}h_{v_{w}}^{L}
\end{equation}
where $v_{w}$ denotes the word nodes and $Y_{w}$ is the clue words probability distribution.

\noindent \textbf{Multi-Task Learning.} In our SHGN model, all three tasks all jointly performed through multi-task learning. The final objective is defined as:

\begin{equation}
\mathcal{L} = \lambda_{1}\mathcal{L}_{sen} + \lambda_{2}\mathcal{L}_{clu} + (1-\lambda_{1}-\lambda_{2})\mathcal{L}_{gen}
\end{equation}
where $\lambda_{1}$ and $\lambda_{2}$ are hyper-parameters.


\section{Experiments}
\label{sec:experiments}

\begin{table}[t]
  \centering
  \caption{Statistics of ROCStories Corpus. Average, minimum, maximum and 95th percentile of length of datasets in wordpieces.}
  \resizebox{0.80\textwidth}{!}
  {
    \begin{tabular}{c|c|cccc|cccc}
    \hline
    \multicolumn{1}{c|}{\multirow{2}{*}{ROCStories Corpus}} & \multicolumn{1}{c|}{\multirow{2}{*}{Samples}} & \multicolumn{4}{c|}{Story Context}                               & \multicolumn{4}{c}{Story Ending}                                \\ \cline{3-10}
    \multicolumn{1}{c|}{}                                   & \multicolumn{1}{c|}{}                         & avg. & min & max & 95ptcl. & avg. & min & max & 95ptcl. \\ \hline
    training                                                & 90,000                                        & 35.0         & 4            & 65             & 48               & 9.5          & 1            & 20             & 14               \\
    validation                                              & 4,081                                         & 32.6         & 15            & 56            & 46               & 8.9          & 2            & 18             & 13               \\
    testing                                                 & 4,081                                         & 33.7         & 12            & 61            & 47               & 9.4          & 2            & 17             & 14               \\ \hline
    \end{tabular}
  }

  \label{table:roc}
\end{table}

\subsection{Experiment Setup}
\noindent\textbf{Dataset.} Following previous work~\cite{Guan2019StoryEG,Huang2021StoryEG}, we evaluate SHGN on the ROCStories Corpus~\cite{Mostafazadeh2016ACA} which contains 98,162 five-sentence daily life stories collected from crowd-workers. Table~\ref{table:roc} shows the detailed statistics of the Corpus. In SEG task, the first four-sentence of each story is regarded as story context while the last sentence is the ground truth ending.

\noindent\textbf{Implementation Details.}
We implement our model based on \texttt{transformers}~\cite{Wolf2019HuggingFacesTS} and \texttt{PyTorch Geometric}~\cite{Fey2019FastGR} libraries. For graph initializer, we utilize the SimCSE~\cite{Gao2021SimCSESC} sentence embedding model released by original authors\footnote{\url{https://huggingface.co/princeton-nlp/sup-simcse-roberta-base}}, which has a similar architecture with \texttt{Robera-base} (768 hidden size, 12 multi-head attention, 12 layers). Following Feng et al.~\cite{Feng2021IncorporatingCK}, the number of graph layer is set to 1 and we set the hidden size to 768. The transformer decoder used in our experiments has 12 decoder layers, 12 multi-head attention and 768 hidden size. To construct labels for sentiment prediction of story endings, we use \texttt{VADER} toolkit\footnote{\url{https://github.com/cjhutto/vaderSentiment}}. The dependency parsing algorithm Biaffine~\cite{Dozat2017DeepBA} used in our experiments is implemented by \texttt{SuPar}\footnote{\url{https://github.com/yzhangcs/parser}} toolkit.

During training, we set the batch size to 64 and use linear warmup of 1,000 steps. We employ grid search of Learning Rate (LR) in [2e-5, 3e-5, 5e-5] and number epochs in [5,10,15]. The best configuration used LR=5e-5, 15 epochs. During inference, we use beam search and the beam size is set to 5.
The coefficient $\lambda_{1}$ and $\lambda_{2}$ used in multi-task learning strategy are both 0.1.

\noindent\textbf{Automatic Evaluation.} We make use of BLEU~\cite{Papineni2002BleuAM} and ROUGE~\cite{Lin2004ROUGEAP} for our automatic evaluation metrics, and report BLEU-1,2,3,4 together with ROUGE-1,2,L scores. Following previous study~\cite{Huang2021StoryEG}, we utilize \texttt{nlg-eval}\footnote{\url{https://github.com/Maluuba/nlg-eval}} and \texttt{pyrouge}\footnote{\url{https://github.com/bheinzerling/pyrouge}} toolkits to calculate the scores. Note that the
BLEU or ROUGE scores might vary with different toolkits.

\noindent\textbf{Human Evaluation.} Considering the limitation of automatic evaluation, it is necessary to conduct human evaluation. Specifically, three aspects are considered as the criteria: (1) Grammaticality evaluates correct, fluent and natural of the generated story endings; (2) Logicality is used to evaluate whether the generated endings are reasonable and coherent; (3) Relevance measures how relevant are the endings to the story context. For a model, We randomly choose 100 generated story endings and employ three NLP postgraduates to make the evaluation. The scoring adopts a 3-point scale, with 1 as the worst and 3 as the maximum.

\subsection{Baseline Methods}
We compare our model with several typical baselines and the state-of-the-art baselines:
\begin{itemize}
\item \textbf{Seq2Seq+Att}~\cite{Luong2015EffectiveAT}: A LSTM-based Seq2Seq model with attention mechanism.
\item \textbf{Transformer}~\cite{Vaswani2017AttentionIA}: Transformers is a parallel Seq2Seq model based on multi-head attention and feed forward networks.
\item \textbf{HLSTM}~\cite{Yang2016HierarchicalAN}: A hierarchical LSTM utilizes word-level and sentence-level LSTM as its encoder, and uses vanilla LSTM as its decoder to generate text sequence.
\item \textbf{IE+MSA}~\cite{Guan2019StoryEG}: A SEG model which considers external commonsense knowledge and uses an incremental encoding scheme to generate endings.
\item \textbf{T-CVAE}~\cite{Wang2019TCVAETC}: A conditional variational auto-encoder model based on transformers.
\item \textbf{Plan\&Write}~\cite{Yao2019PlanAndWriteTB}: A story generation model, which first uses a given title (topic) to obtain several keywords, and then generates complete stories.
\item \textbf{MGCN-DP}~\cite{Huang2021StoryEG}: The state-of-the-art SEG model which utilizes the dependency parse tree to construct a graph for story context, and makes use of GCN to capture story clues. A transformer decoder is employed to generate final endings.
\end{itemize}


\subsection{Main Results}
Table~\ref{table:results} shows the results of automatic evaluation. The results show that our model significantly outperforms these baselines. Specifically, our model achieves an improvement of 5.3\% / 5.3\% / 4.9\% / 4.5\% over the IE+MSA / T-CVAE / Plan\&Write / MGCN-DP in term of B1. As for B2, our model outperforms the IE+MSA / T-CVAE / Plan\&Write / MGCN-DP by 20.5\% / 22.1\% / 11.9\% / 8.0\%, respectively. With respect to B4, our model implements an improvement of 28.6\% / 35.0\% / 17.4\% / 8.0\%. And for R-L, our model achieves an improvement of 13.0\% / 13.0\% / 9.8\% / 7.3\%.
Other automatic metrics (i.e., B3, R1 and R2) also demonstrate the superiority of our SHGN.
The results indicate that our model can comprehend the story context better based on the heterogeneous graph which model both the information of story context at different granularity (knowledge, sentence and word) levels and the multi-grained interactive relations among them.

\begin{table*}[t]
  \centering
  \caption{Experiments on the ROCStories Corpus for the SEG task. The \textbf{bold} denote the best performance. For performances of baseline methods, $^{\dagger}$ represents the reproducing results while $^{\ddagger}$ denotes the results reported by Huang et al.~\cite{Huang2021StoryEG}.}
  \resizebox{0.80\textwidth}{!}
  {
    \begin{tabular}{lcclllll}
    \hline
    Model       & B1   & B2  & B3 & B4 & R1 & R2 & R-L \\ \hline
    Seq2Seq+Att~\cite{Luong2015EffectiveAT} &\ 18.5$^{\ddagger}$\  &\ 5.9$^{\ddagger}$\ &\ -\ &\ -\ &\ -\ &\ -\ &\ -\ \\
    Transformer~\cite{Vaswani2017AttentionIA} &\ 17.4$^{\ddagger}$\ &\ 6.0$^{\ddagger}$\ &\ -\ &\ -\ &\ -\ &\ -\ &\ -\ \\
    HLSTM~\cite{Yang2016HierarchicalAN}       &\ 22.1$^{\ddagger}$\ &\ 7.1$^{\ddagger}$\ &\ -\ &\ -\ &\ -\ &\ -\ &\ -\ \\
    IE+MSA~\cite{Guan2019StoryEG}      &\ 24.3$^{\dagger}$\ &\ 7.8$^{\dagger}$\ &\ 3.9$^{\dagger}$\ &\ 2.1$^{\dagger}$\ &\ 17.5$^{\dagger}$\ &\ 2.9$^{\dagger}$\ &\ 20.8$^{\dagger}$\ \\
    T-CVAE~\cite{Wang2019TCVAETC}      &\ 24.3$^{\dagger}$\ &\ 7.7$^{\dagger}$\ &\ 3.8$^{\dagger}$\ &\ 2.0$^{\dagger}$\ &\ 17.6$^{\dagger}$\ &\ 3.0$^{\dagger}$\ &\ 20.8$^{\dagger}$\ \\
    Plan\&Write~\cite{Yao2019PlanAndWriteTB} &\ 24.4$^{\dagger}$\ &\ 8.4$^{\dagger}$\ &\ 4.1$^{\dagger}$\ &\ 2.3$^{\dagger}$\ &\ 18.1$^{\dagger}$\ &\ 3.3$^{\dagger}$\ &\ 21.4$^{\dagger}$\ \\
    MGCN-DP~\cite{Huang2021StoryEG}     &\ 24.5$^{\dagger}$ (24.6$^{\ddagger}$)\ &\ 8.7$^{\dagger}$ (8.6$^{\ddagger}$)\ &\ 4.3$^{\dagger}$\ &\ 2.5$^{\dagger}$\ &\ 18.4$^{\dagger}$\ &\ 3.5$^{\dagger}$\ &\ 21.9$^{\dagger}$\ \\ \hline
    SHGN(Our) &\ \textbf{25.6}\ &\ \textbf{9.4}\ &\ \textbf{4.7}\ &\ \textbf{2.7}\ &\ \textbf{20.3}\ &\ \textbf{3.9}\ &\ \textbf{23.5}\ \\ \hline
    \end{tabular}
  }
  \label{table:results}
\end{table*}

\begin{table*}[t]
  \centering
  \caption{ROCStories Corpus ablations. ``w/o'' means ``without''. glob.: global, know.: knowledge, init.: initializer, SESP.: sentiment prediction of story endings, CWP.: clue words prediction. The \textbf{bold} and \uline{underline} denote the best and the second performances, respectively.}
  \resizebox{0.50\textwidth}{!}
  {
    \begin{tabular}{c|lccc}
    \cline{1-5}
    \#\ & Model           & B2           & B4           & R-L \\ \cline{1-5}
    1\ \ & SHGN            &\ \textbf{9.4}\ &\ \textbf{2.7}\ &\ \textbf{23.5}\ \\ \cline{1-5}
    2\ \ & SHGN(w/o glob.) &\ 9.2\  &\ \textbf{2.7}\ &\ 22.7\ \\
    3\ \ & SHGN(w/o know.) &\ 9.0\  &\ 2.6\ &\ 22.4\ \\
    4\ \ & SHGN(w/o word)  &\ 8.9\ &\ 2.6\ &\ 22.3\ \\ \cline{1-5}
    5\ \ & SHGN(init. Sentence-BERT)  &\ \uline{9.3}\ &\ \textbf{2.7}\ &\ \uline{23.2}\ \\
    6\ \ & SHGN(init. LSTM)  &\ 7.6\ &\ 2.3\  &\ 21.2\ \\ \cline{1-5}
    7\ \ & SHGN(w/o mutli-task)  &\ 9.1\ &\ 2.5\ &\ 22.7\ \\
    8\ \ & SHGN(w/o SPSE.)  &\ \uline{9.3}\ &\ 2.6\ &\ 23.1\ \\
    9\ \ & SHGN(w/o CWP.)  &\ 9.2\ &\ 2.5\ &\ 22.9\ \\ \cline{1-5}
    \end{tabular}
  }
  \label{table:ablation}
\end{table*}

\subsection{Ablation Study}
\label{sec:ablations}

\noindent\textbf{Effectiveness of Heterogeneous Graph.} As described in Section~\ref{subsec:graph_construct}, the constructed heterogeneous story graph contains four types of nodes: \textit{global}, \textit{knowledge}, \textit{sentence} and \textit{word}. We run 3 ablations, modifying various settings of our SHGN: (1) remove global node; (2) remove knowledge nodes; (3) remove word nodes. The effect of these ablations is shown in Table~\ref{table:ablation} (row 1 vs. row 2-4). In each case, the automatic evaluation scores are lower than our origin SHGN, which justifies the rationality of our model.

\noindent\textbf{Effectiveness of Graph Initializer.} We utilize SimCSE as our graph initalizer. We also run 2 ablations: (1) replace SimCSE~\cite{Gao2021SimCSESC} with Sentence-BERT~\cite{reimers-gurevych-2019-sentence}; (2) replace SimCSE with BiLSTM, where the forward and backward hidden states are concatenated as the initial node representations. GloVe.6B~\cite{pennington-etal-2014-glove} word embedding (300 dimension) is used in the BiLSTM initializer. Table~\ref{table:ablation} (row 1 vs. row 5, 6) shows the effectiveness of SimCSE.
Specifically, the results of BiLSTM initializer are dramatically dropped compared to pretrained sentence embedding models, which indicates the superiority of the pretrained models.

\noindent\textbf{Effectiveness of Multi-Task Learning.} In order to demonstrate the effectiveness of our auxiliary tasks, we remove each of them and all of them in ablation studies, respectively. The results are shown in Table~\ref{table:ablation} (row 1 vs. row 7-9), which indicates our designed auxiliary tasks can facilitate story context comprehension.


\begin{figure}[t]
\centering
\subfigure[Grammaticality]{
  \includegraphics[width=0.25\linewidth]{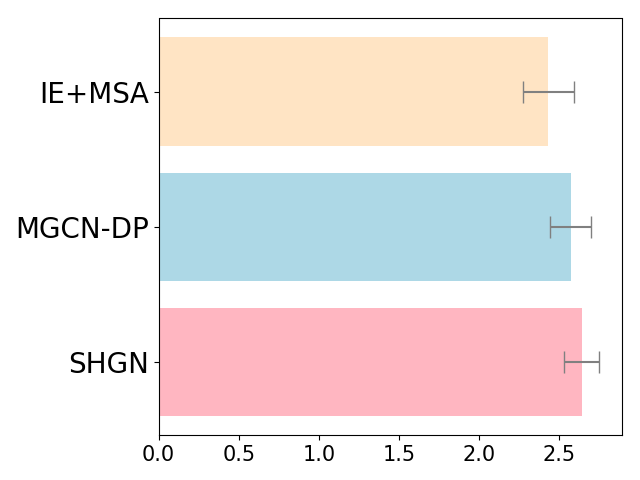}
}
\subfigure[Logicality]{
  \includegraphics[width=0.25\linewidth]{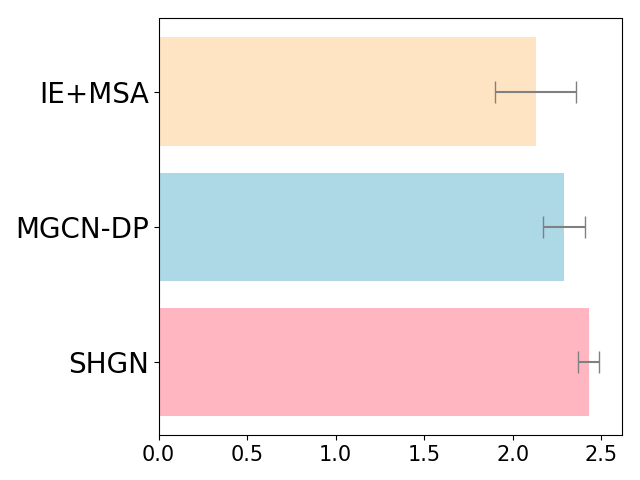}
}
\subfigure[Relevance]{
  \includegraphics[width=0.25\linewidth]{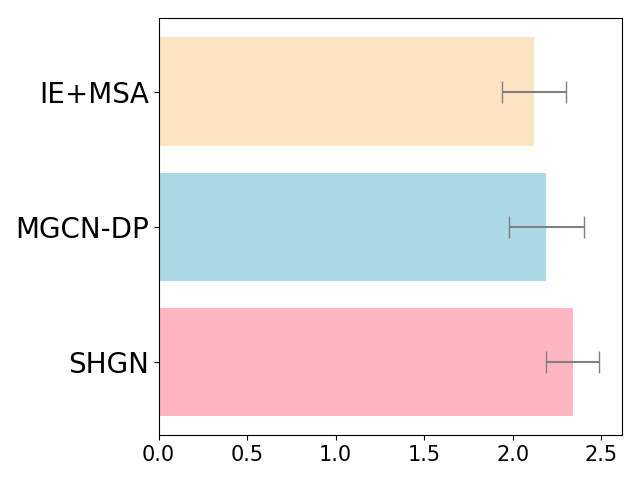}
}
\caption{Results on human evaluation, including means and variances. Our SHGN model outperforms IE+MSA and MGCN-DP on all three aspects.}
\label{fig:huamn_evaluation}
\end{figure}


\subsection{Human Study and Case Study}
We conduct human study on IE+MSA~\cite{Guan2019StoryEG}, MGCN-DP~\cite{Huang2021StoryEG} and our SHGN. Figure~\ref{fig:huamn_evaluation} shows the results of human study. Our SHGN performs better than IE+MSA and MGCN-DP on all three aspects, which verifies that our SHGN performances better on generating reasonable and coherent story endings.

To provide a deeper understanding of generated endings, we show two examples from different models in Table~\ref{table:case_study}.
In Case 1, the IE+MSA model performs worst, since it generates a illogical story ending. In detail, the ``\textit{new job}'' is irrelevant to the context.
The MGCN-DP and SHGN output the endings with content about ``\textit{stomach problems}'', which more relevant to the context, but they miss the context clue ``\textit{couldn't figure out}''. This finding shows that: (1) The graph architecture could model the rich interactive relations across story and improve the relevance of generated endings; (2) SEG is still a challenging task.
In Case 2, we know that there may be a bad ending for Keith since he did something illegal at work.
However, MGCN-DP generates an unreasonable ending, which is contrary to the gold ending. IE+MSA and SHGN effectively capture the context clues and generate bad endings. Furthermore, the generated ending of SHGN is highly consistent with the gold ending, which shows the superiority of our model.

\begin{table*}[t]
  \centering
  \caption{Case Study on ROCStories Corpus. \textbf{Bold} words represent the key entities, events, or key words. \textit{italic} words denote improper words.}
  \resizebox{0.95\textwidth}{!}
  {
    \begin{tabular}{l|l|l}
    \hline
        & \multicolumn{1}{c|}{Case 1} & \multicolumn{1}{c}{Case 2} \\
    \hline
    Context     & \tabincell{l}{Tim always had \textbf{stomach problems}. 
    \\ He tried different things to \textbf{fix them}. 
    \\ His doctors \textbf{couldn't} really \textbf{figure out} \\ 
    what was wrong.  \\ 
    It really \textbf{cut into} Tim's social \textbf{life}.} & \tabincell{l}{Keith was \textbf{working at} a mechanic \textbf{shop}. \\ 
    He had \textbf{given} a customer a \textbf{high quote}. \\ 
    Keith \textbf{kept the difference} between the \\ 
    quote and the actual bill. Keith's boss  \\ 
    \textbf{found out} what \textbf{he did}. } \\ \hline
    IE+MSA      & Tim was able to get a \textit{new job}.  & Keith's boss was \textbf{mad} at him.                                                                                                                                                                                     \\
    MGCN-DP     & Tim \textbf{felt better} after that.                                                                                                                                          & Keith \textit{got} the \textbf{job}.                                                                                                                                                                  \\
    SHGN        & Tim eventually \textbf{got better}                                                                                                                                                & Keith was \textbf{fired} from his \textbf{job}.                                                                                                                                                        \\ \hline
    Gold & \tabincell{l}{Eventually he learned to just \textbf{live with} \\ the \textbf{discomfort}.  }                                                                                                                 & Keith \textbf{lost} his \textbf{job}.                                                                                                                                                                 \\ \hline
    \end{tabular}
  }
  \label{table:case_study}
\end{table*}

\section{Conclusion}
\label{sec:conclusion}
In this paper, we study SEG and propose Story Heterogeneous Graph Network (SHGN) which utilizes graph architecture and commonsense knowledge to comprehend the story context and handle the implicit knowledge behind the story. Besides, two auxiliary tasks are introduced to facilitate story comprehension. Extensive experiments on widely used ROCStories Corpus show that our SHGN achieves new state-of-the-art performances. Human study and case study further prove the effectiveness of our model.

\section*{Acknowledgement}
%
This research is supported by the National Natural Science Foundation of China (Grant No. 62072323, 62102276), the Natural Science Foundation of Jiangsu Province (No. BK20191420, BK20210705, BK20211307), the Major Program of Natural Science Foundation of Educational Commission of Jiangsu Province, China (Grant No. 19KJA610002, 21KJD520005), the Priority Academic Program Development of Jiangsu Higher Education Institutions, and the Collaborative Innovation Center of Novel Software Technology and Industrialization.
We would like to thank Duo Zheng and the members of the NLPTTG group, Netease Fuxi AI Lab for the helpful discussions and valuable feedback. We also thank anonymous reviewers for their suggestions and comments.



\bibliographystyle{splncs04}
\bibliography{references}

\end{document}